\documentclass[10pt]{article}

\usepackage{graphicx}
\usepackage{listings}
\usepackage{xcolor} 
\usepackage{url}
\usepackage{natbib}
\usepackage{authblk}
\usepackage[colorlinks=true, citecolor=black, urlcolor=black, linkcolor=black]{hyperref} %
\usepackage{cleveref}
\usepackage{booktabs}
\usepackage[margin=1in]{geometry}
\usepackage{subcaption} %
\usepackage[whole]{bxcjkjatype}
\usepackage{makecell}
\usepackage{todonotes}

\title{PLaMo 2 Technical Report}
\author{Preferred Networks, Inc.}

\affil{\texttt {plamo2-tech-report@preferred.jp}}

\date{}

\begin{document}

\maketitle

\begin{abstract}
In this report, we introduce PLaMo 2, a series of Japanese-focused large language models featuring a hybrid Samba-based architecture that transitions to full attention via continual pre-training to support 32K token contexts. Training leverages extensive synthetic corpora to overcome data scarcity, while computational efficiency is achieved through weight reuse and structured pruning. This efficient pruning methodology produces an 8B model that achieves performance comparable to our previous 100B model. Post-training further refines the models using a pipeline of supervised fine-tuning (SFT) and direct preference optimization (DPO), enhanced by synthetic Japanese instruction data and model merging techniques. Optimized for inference using vLLM and quantization with minimal accuracy loss, the PLaMo 2 models achieve state-of-the-art results on Japanese benchmarks, outperforming similarly-sized open models in instruction-following, language fluency, and Japanese-specific knowledge.
\end{abstract}

\section{Introduction}

The landscape of large language models (LLMs) has evolved rapidly, with new architectures and training methodologies continuously pushing the boundaries of performance. However, several key challenges persist: the immense computational cost of training, the difficulty of achieving efficient inference for long sequences, and the scarcity of high-quality, large-scale training data for languages other than English. These challenges are particularly acute for developing high-performance models specialized in languages, such as Japanese.

This technical report introduces PLaMo 2, a new series of LLMs developed by Preferred Networks, based on the foundations of our previous PLaMo-100B model \citep{plamo100b}.
The primary goal of PLaMo 2 is to deliver state-of-the-art performance, especially in the Japanese language, while simultaneously addressing the challenges of training efficiency and long-context capability. To achieve this, we introduce a multifaceted approach that includes innovations in model architecture, data generation, and training strategies.

Our key contributions are threefold.
First, we adopt a hybrid architecture based on Samba \citep{samba}, combining Mamba \citep{mamba} with sliding window attention \citep{Beltagy2020arXiv}, which is later evolved to full attention during a continual pre-training phase to overcome long-context retrieval limitations as discussed in \Cref{subsec:cpt}.
Second, we significantly augment our training data through the extensive use of high-quality synthetic data generated via LLMs, focusing on translation, paraphrasing, code, and mathematics to compensate for the limited availability of data resources.
Third, it employs an efficient training paradigm. Specifically, it applies weight reusing for initializing larger models and utilizes pruning techniques to generate high-performance, smaller models from larger parent models.
This allows us to develop a family of models at different scales efficiently.

Our comprehensive evaluations demonstrate that the PLaMo 2 models, particularly PLaMo 2.1-8B and PLaMo 2.0-31B, achieve state-of-the-art results on a wide array of Japanese benchmarks.
Notably, our 8B model achieves performance comparable to or even exceeding our previous 100B-parameter model, showcasing the effectiveness of our methods. This paper details the architectural choices, data curation strategies, and training procedures for both pre-training and post-training, followed by a thorough analysis of benchmark results that underscore the capabilities of the PLaMo 2 series.

\section{Pre-Training}

\subsection{Model Architecture}
\label{subsec:model_arch}
The architecture of LLMs plays a crucial role in determining various key capabilities, including training stability, efficiency, and performance during training and inference. Numerous architectural modifications have been proposed, such as RMSNorm \citep{rms_norm} and Grouped Query Attention \citep{gqa}, which have been widely adopted.

In our previous version, PLaMo-100B \citep{plamo100b}, we employed a Transformer \citep{transformer}-based architecture. This choice was motivated by the fact that, as of 2024, most open-source LLMs still use Transformer architecture, which remains the most successful approach in terms of training speed and achievable performance.
However, Transformer architecture is known to exhibit limitations in inference performance when dealing with long sequence lengths. Two particular challenges stand out: first, inference beyond the maximum sequence length during training requires specialized techniques; second, longer sequence lengths inevitably lead to increased memory consumption that prevents inference altogether or significantly slows it down.

\begin{figure}[t]
    \centering
\includegraphics[width=0.5\textwidth]{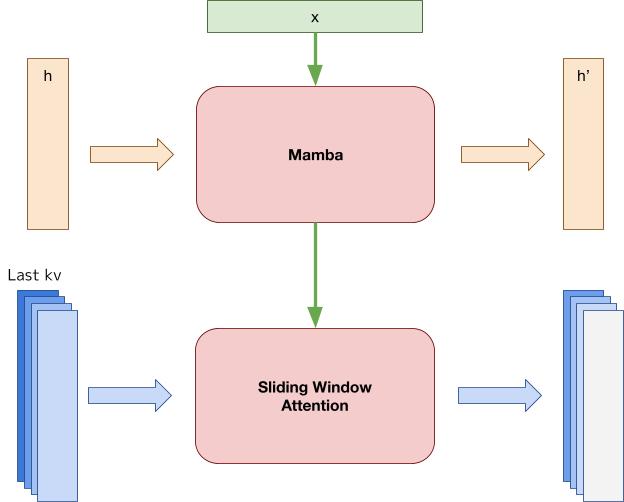}
    \caption{Samba architecture, combining Mamba with sliding window attention.
    Note that $\mathbf{x}$ is a sequence of token features, $\mathbf{h}$ and $\mathbf{h}'$ are mamba's hidden states, and \textit{kv} stands for the key-value cache.}
    \label{fig:samba}
\end{figure}

To address these issues, we implemented Samba-based architecture \citep{samba} to address primarily the second of these challenges.
\Cref{fig:samba} depicts the overview of our architecture.
Samba combines Mamba \citep{mamba} with sliding window attention \citep{Beltagy2020arXiv} and features the following key characteristics:
\begin{enumerate}
    \item Enables seamless inference and generation even for sequence lengths exceeding those used during training.
    \item Maintain constant memory usage and computational complexity regardless of the length of the sequence.
\end{enumerate}

\subsection{Training Data}
For PLaMo 2, we have expanded our dataset resources beyond those previously used for PLaMo-100B by developing synthetic datasets through various methods to enhance model performance.
We will now introduce the representative datasets used in our training process.

\subsubsection{Web Corpus}
To acquire general Japanese and English knowledge, we use a web corpus.
In both languages, data are extracted from CommonCrawl, with English using primarily an open dataset while the Japanese dataset was derived from PLaMo-100B through processes including filtering and duplicate removal.

Regarding the filtering process of the Japanese dataset, we developed a new dataset by applying the following filters based on English research, such as FineWeb-Edu \citep{fineweb} and DataComp-LM \citep{dclm}:
\begin{enumerate}
    \item Educational value-based filtering similar to FineWeb-Edu
    \item Categorization and downsampling of web data
\end{enumerate}
For both approaches, we first created a lightweight model from instruction-tuned PLaMo-100B annotations, then performed filtering using this lightweight model before incorporating the results for PLaMo 2's training.

Through method 1, which relies on educational value-based datasets, we aim to reduce unexpected performance degradation.
Method 2 seeks to compensate for any knowledge loss from method 1 by supplementing it with additional data.

\subsubsection{Web Coding Data}
To improve coding performance, we used existing datasets publicly available on platforms like Hugging Face as well as data extracted from the web.

For web data extraction, we followed a similar preprocessing approach to our Japanese dataset for PLaMo-100B, extracting data deemed particularly relevant to coding from CommonCrawl data.
This time, we implemented the following methods:
\begin{enumerate}
    \item Removal of irrelevant data through filtering: Since parsing all HTML content from CommonCrawl data to extract text would be computationally expensive, we applied simple filters to remove data that are unlikely to contain coding content. Specifically, we excluded data lacking common coding tags such as \textless pre\textgreater \ and \textless code\textgreater.
    \item Conversion of filtered data into a markdown format
    \item Classification of data as coding content using a fastText model \citep{Joulin2017EACL}: This model was trained on data labeled with an LLM.
\end{enumerate}

\subsubsection{Synthetic Data}
The training of many LLMs, including Phi-4, has increasingly incorporated synthetic data generated through LLMs and other methods.
For the development of PLaMo 2, we primarily used synthetic data to enhance our high-quality Japanese dataset, focusing on translation, paraphrasing, coding, and mathematical tasks. We will explain each of these approaches in turn.\\

\noindent\textbf{Translation/Paraphrasing}\ 
While high-quality Japanese data exists, such as Wikipedia, the dataset size remains significantly smaller than that of English. To address this challenge, we used two key methodologies:
\begin{itemize}
    \item Generating high-quality Japanese data by translating high-quality English data using an LLM
    \item Creating paraphrased versions of high-quality Japanese data
\end{itemize}

For paraphrasing, we adapted existing techniques \citep{rephrasing_the_web} and developed multiple LLM-based prompt systems, including those for style conversion and QA-style conversion. Additionally, we implemented rule-based paraphrasing using Self-Tuning \citep{self_tuning} approaches.
The source data for translation tasks included multiple high-quality English datasets, such as Wikipedia.
For paraphrasing, we specifically processed Japanese Wikipedia data.
We used PLaMo-100B, an LLM renowned for its exceptional Japanese language capabilities, for both translation and paraphrasing tasks.\\

\noindent\textbf{Code Data Generation Using Existing Data}\
We generated new data by processing existing code datasets through LLMs. Specifically, we employed LLMs to add comments to code and to generate paraphrased versions while preserving the original code's meaning as much as possible.

\begin{lstlisting}
import argparse
 
parser = argparse.ArgumentParser()
parser.add_argument("input", type=str)
parser.add_argument("--first-n-lines", type=int)
args = parser.parse_args()
 
with open(args.input, "r") as f:
    lines = []
    for line in f.readlines():
        lines.append(line.rstrip())
 
if args.first_n_lines:
    lines = lines[:args.first_n_lines]
 
lines.sort()
print("\n".join(lines))
\end{lstlisting}

By providing this code to the LLM, we obtain the following descriptive text:
\texttt{
    This Python script reads a file specified by the user, optionally limits the number of lines read, sorts those lines, and prints them. It uses the \texttt{argparse} module to handle command-line arguments for the input file and the number of lines to consider. The script reads the file, strips newline characters from each line, optionally truncates the list to the specified number of lines, sorts the remaining lines, and prints them joined by newline characters.
}

By re-generating code from this description, we can obtain the following code. This approach enables us to generate code data that is more understandable and useful for learning purposes.

\begin{lstlisting}
import argparse
 
def main():
    # Set up argument parser
    parser = argparse.ArgumentParser(description='Read a file, optionally ...')
    parser.add_argument('file', type=str, help='The input file to read')
    parser.add_argument('-n', '--lines', type=int, default=None, ...
 
    # Parse command-line arguments
    args = parser.parse_args()
 
    try:
        # Read the file
        with open(args.file, 'r', encoding='utf-8') as file:
            lines = file.readlines()
 
        # Strip newline characters
        lines = [line.rstrip('\n') for line in lines]
 
        # Optionally limit the number of lines
        if args.lines is not None:
            lines = lines[:args.lines]
 
        # Sort the lines
        lines.sort()
 
        # Print the sorted lines joined by newline characters
        print('\n'.join(lines))
 
    except FileNotFoundError:
        print(f"Error: The file '{args.file}' was not found.")
    except Exception as e:
        print(f"An error occurred: {e}")
 
if __name__ == '__main__':
    main()
\end{lstlisting}

\noindent\textbf{Code Data Generation Based on Prompts}\
As indicated in its name, Phi generates textbook-style data for training purposes to enhance LLM performance. We similarly attempted to produce large amounts of high-quality data.
Our approach mainly followed the Magicoder methodology \citep{magicoder}.

The Magicoder technique incorporates code snippets within specific portions of the prompt to enable diverse data generation. While Magicoder primarily creates instruction data consisting of problem-solution pairs, we also modified the prompt to generate code data similar to textbook examples and standard software code.
Furthermore, we included randomly selected "topic" words in the prompt to facilitate data generation in a wider variety of contexts.
We also experimented with generating more complex data by having the model produce additional data based on the initially generated material.
Additionally, Nemotron-4 \citep{Nvidia2024arXiv} proposes a method that uses LLMs to identify key topics related to a specific theme (e.g., for ``learning Python,'' topics might include syntax and libraries). We incorporated similar data generation approaches in some of our experiments.

\noindent\textbf{Math Data Generation}\
For generating synthetic math data, we used existing mathematical datasets as seed data in conjunction with LLMs.
We employed GSM8k \citep{Cobbe2021arXiv} and Lila \citep{Mishra2022EMNLP} as our seed datasets.
The data generation process was implemented in two stages:
\begin{enumerate}
    \item Generating new problems from the seed data
    \item Generating both solutions and the reasoning processes leading to those solutions
\end{enumerate}
For this problem generation and other related tasks, we primarily used Phi-3.5.

\subsection{Leveraging Pre-trained Weights}

The training process for PLaMo 2 incorporated pre-trained model weights to minimize training token requirements while achieving performance improvements.
We employed two complementary approaches: weight reusing to initialize larger models using weights from smaller models, and pruning to use larger models for training smaller models.

\subsubsection{Weight Reusing}

Weight reusing is a technique where the initial weights of a larger DNN model are determined using weights from a smaller DNN model. In the context of LLMs, this method was first introduced in the NeurIPS presentation of Phi-2 \citep{phi2}.

\begin{figure}[h]
    \centering
\includegraphics[width=0.5\textwidth]{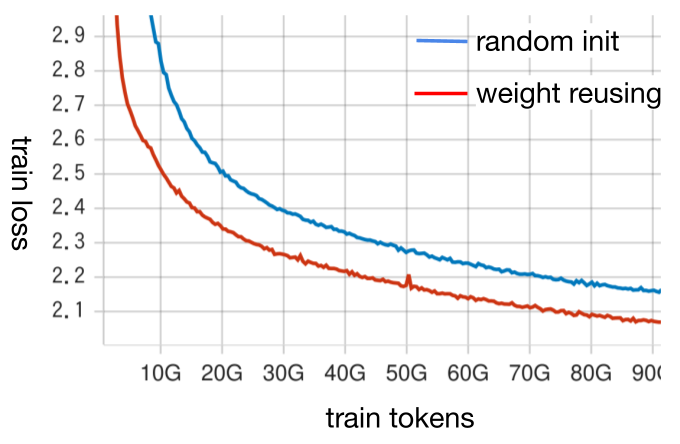}
    \caption{Training loss during weight reusing experiments. Blue represents training an 8B model from random initialization, while red shows results using weight reusing initialization.}
    \label{fig:loss-weight-reusing}
\end{figure}

Preliminary experiments with smaller models demonstrated significant performance gains through weight reusing.
The initial training phase showed a marked improvement in train loss reduction, and throughout subsequent training stages, the model maintained a lower train loss compared to random initialization, as shown in \Cref{fig:loss-weight-reusing}.
Performance improvements were also observed in various downstream tasks.

Typically, such techniques tend to show diminishing or negative effects as training progresses.
However, given the enormous computational resources required for extended LLM training, we anticipate that weight reusing will generally yield more beneficial results than detrimental ones.

\subsubsection{Pruning}

Pruning is a method that retains only the most critical weights from a pre-trained DNN model, enabling efficient creation of high-performance, compact DNN models. In LLMs, Llama 3.2 \citep{llama3_2} employed pruning to develop a 1B-scale model. We developed the 8B parameter PLaMo 2.1 8B model from PLaMo 2 31B through this pruning process.

We implemented a pruning approach based on the Minitron \citep{minitron} methodology, which combines structural pruning with retraining. We selected this method due to its relatively lightweight pruning process and the advantage of maintaining the original model architecture, making the resulting models more practical for deployment.

\begin{figure}[h]
    \centering
\includegraphics[width=0.5\textwidth]{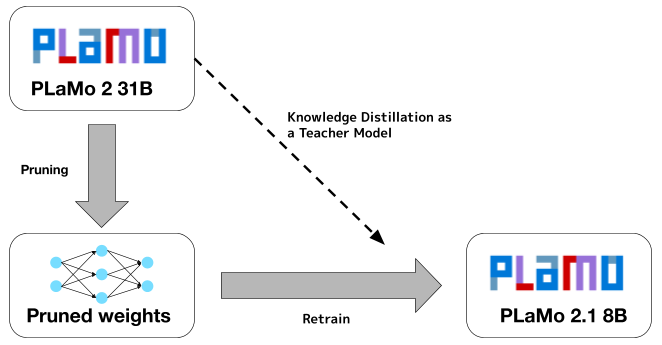}
    \caption{Overview of the pruning process. After pruning PLaMo 2 8B to retain only essential weights, we performed retraining using knowledge distillation as the teacher model.}
\end{figure}

This approach involves first performing pruning to retain only crucial weights from the original model, followed by retraining using knowledge distillation from the original model as the teacher.
While we employed the standard KL divergence-based loss function for knowledge distillation, to reduce memory consumption, we limited the teacher model's logits to the top 128 tokens.
Although PLaMo 2's tokenizer supports a vocabulary of 100,000 tokens, storing all logits would result in extremely high memory usage.
By restricting to just the top 128 tokens, we achieved significant memory reduction.
There is a risk that the effectiveness of knowledge distillation diminishes due to potentially inaccurate teacher logits, but we confirmed that using only the top 128 logits does not significantly impact performance, making memory efficiency our primary consideration in our preliminary experiments.

\subsection{Experiment}

\subsubsection{Comparison Models}
For comparison, we evaluated the following models trained on Japanese-included datasets at comparable pretraining levels:
\begin{itemize}
    \item Qwen/Qwen2.5-7B \citep{qwen2.5}
    \item Qwen/Qwen2.5-32B \citep{qwen2.5}
    \item Qwen/Qwen3-8B-Base \citep{qwen3technicalreport}
    \item tokyotech-llm/Llama-3.1-Swallow-8B-v0.2 \citep{Fujii:COLM2024,Okazaki:COLM2024}
    \item sbintuitions/sarashina2-7b \citep{sharashina2-7b}
    \item abeja/ABEJA-Qwen2.5-32b-Japanese-v0.1 \citep{abeja}
    \item pfnet/plamo-100b (PLaMo-100B)
\end{itemize}
Furthermore, we included for evaluation the PLaMo 2.0 8B/31B models trained without pruning or knowledge distillation, as well as the PLaMo 2.1 8B model developed using both pruning and knowledge distillation techniques.

\subsubsection{Benchmarks}

\noindent\textbf{JMMLU, MMLU}\
For JMMLU \citep{mmlu}, we randomly selected five sample questions from each test dataset and used them as few-shot prompts. Evaluation was conducted using modified versions of the MMLU code from the Language Model Evaluation Harness \citep{eval-harness}.
Regarding multiple-choice questions like those in MMLU \citep{mmlu}, we observed that prompt formulation significantly impacts performance. We employed the default prompt format from the Language Model Evaluation Harness' MMLU implementation, which involves presenting the options to the LLM and having it select A, B, C, or D as the answer.

\noindent\textbf{JHumanEval, HumanEval+}\
For coding benchmark evaluations, we used HumanEval+ \citep{evalplus} and JHumanEval \citep{jhumaneval}.

\noindent\textbf{pfgen-bench}\
pfgen-bench \citep{pfgen} is a Japanese text generation performance evaluation benchmark. The benchmark scores range from 0 to 1, with higher values indicating superior Japanese text generation quality.

\noindent\textbf{wmt20}\
For the wmt20 \citep{wmt20} evaluation, we modified Stability AI's lm-evaluation-harness \citep{stability_lm_eval} and used wmt22-comet-da \citep{comet} as the evaluation metric.
This wmt22-comet-da metric also ranges from 0 to 1, with higher values indicating better translation.

\subsubsection{Benchmark Results}
\Cref{tab:MMLU,tab:humaneval,tab:pfgen-pretrain,tab:wmt20} show the results of these benchmarks.

\begin{table}[t]
    \centering
    \caption{JMMLU and MMLU Results}
    \label{tab:MMLU}
    \begin{tabular}{lrr}
        \toprule
        Model   & JMMLU (5 shot, acc) & MMLU (5 shot, acc) \\
        \midrule
        Qwen/Qwen2.5-7B & 	0.681       & 0.742     \\
        Qwen/Qwen3-8B-Base & 0.714       & 0.765     \\
        tokyotech-llm/Llama-3.1-Swallow-8B-v0.2 & 0.600        & 0.622 \\
        sbintuitions/sarashina2-7b & 0.400 & 0.425 \\
        PLaMo 2 8B & 0.572 & 0.573 \\
        PLaMo 2.1 8B & 0.635 & 0.635 \\
        \midrule
        Qwen/Qwen2.5-32B & 0.786 & 0.832 \\
        abeja/ABEJA-Qwen2.5-32b-Japanese-v0.1 & 0.802 & 0.826 \\
        PLaMo 2 31B & 0.672 & 0.681 \\
        \midrule
        pfnet/plamo-100b & 	0.575 & 0.603 \\
        \bottomrule
    \end{tabular}
\end{table}

\begin{table}[t]
    \centering
    \caption{JHumanEval and HumanEval+ Results}
    \label{tab:humaneval}
    \begin{tabular}{lrr}
        \toprule
        Model   & JHumanEval (0-shot, pass@1) & HumanEval+ (0-shot, pass@1) \\
        \midrule
        Qwen/Qwen2.5-7B & 0.482       & 0.470     \\
        Qwen/Qwen3-8B-Base &    	0.604    & 	0.585     \\
        tokyotech-llm/Llama-3.1-Swallow-8B-v0.2 & 0.232       & 0.201 \\
        sbintuitions/sarashina2-7b & 0.128 & 0.128 \\
        PLaMo 2 8B & 0.463 & 0.463 \\
        PLaMo 2.1 8B & 0.372 & 	0.463 \\
        \midrule
        Qwen/Qwen2.5-32B & 	0.628 & 0.463 \\
        abeja/ABEJA-Qwen2.5-32b-Japanese-v0.1 & 0.463 & 0.213 \\
        PLaMo 2 31B & 	0.488 & 	0.555 \\
        \midrule
        pfnet/plamo-100b & 	0.268 & 0.220 \\
        \bottomrule
    \end{tabular}
\end{table}

\begin{table}[t]
    \centering
    \caption{pfgen-bench Results}
    \label{tab:pfgen-pretrain}
    \begin{tabular}{lrr}
        \toprule
        Model   & pfgen-bench \\
        \midrule
        Qwen/Qwen2.5-7B &  0.467   \\
        Qwen/Qwen3-8B-Base    &  0.560     \\
        tokyotech-llm/Llama-3.1-Swallow-8B-v0.2     & 0.702  \\
        sbintuitions/sarashina2-7b  & 0.646 \\
        PLaMo 2 8B  & 0.753 \\
        PLaMo 2.1 8B  & 0.725 \\
        \midrule
        Qwen/Qwen2.5-32B  & 0.587 \\
        abeja/ABEJA-Qwen2.5-32b-Japanese-v0.1  & 0.774 \\
        PLaMo 2 31B  & 	0.817 \\
        \midrule
        pfnet/plamo-100b  & 0.747 \\
        \bottomrule
    \end{tabular}
\end{table}

\begin{table}[t]
    \centering
    \caption{WMT20 Results}
    \label{tab:wmt20}
    \begin{tabular}{lrr}
        \toprule
        Model   & \makecell{WMT20 (4-shot, \\ ja $\rightarrow$ en, wmt22-comet-da)} & \makecell{WMT20 (4-shot, \\ja $\rightarrow$ en, wmt22-comet-da)} \\
        \midrule
        Qwen/Qwen2.5-7B & 	0.868       & 0.806    \\
        Qwen/Qwen3-8B-Base &    	0.879	&  0.815  \\
        tokyotech-llm/Llama-3.1-Swallow-8B-v0.2 & 0.900    & 	0.820 \\
        sbintuitions/sarashina2-7b & 0.891 & 0.805 \\
        PLaMo 2 8B & 0.901 & 0.814 \\
        PLaMo 2.1 8B & 0.905 & 	0.821 \\
        \midrule
        Qwen/Qwen2.5-32B & 	0.895 & 0.824 \\
        abeja/ABEJA-Qwen2.5-32b-Japanese-v0.1 & 0.903 & 0.826 \\
        PLaMo 2 31B & 	0.907 & 0.826 \\
        \midrule
        pfnet/plamo-100b & 	0.899 & 0.819 \\
        \bottomrule
    \end{tabular}
\end{table}

Comparing PLaMo 2.1 8B's performance on JMMLU and MMLU datasets, it achieves higher scores than sarashin2-7b and tokyotech-llm/Llama-3.1-Swallow-8B-v0.2—both Japanese-developed models at comparable sizes.
In addition, it matches nearly the same performance level as PLaMo-100B.
We attribute this achievement to the fact that the training data we created since developing PLaMo-100B was of higher quality.

When comparing Qwen/Qwen3-8B-Base with PLaMo 2.1 8B, Qwen/Qwen3-8B-Base outperforms PLaMo 2.1 8B not only in MMLU but also in JMMLU.
This suggests that for knowledge domains required in JMMLU, the linguistic dependency is relatively low, allowing Qwen/Qwen3-8B-Base—which achieves high scores in English and Chinese—to maintain strong accuracy.

The code generation performance follows a similar pattern to MMLU and JMMLU. When comparing Qwen/Qwen3-8B-Base with PLaMo 2.1 8B, Qwen/Qwen3-8B-Base outperforms not only in HumanEval+ but also in JHumanEval. This indicates that code generation capabilities are similarly language-independent, allowing Qwen/Qwen3-8B-Base—which excels in English and Chinese—to achieve superior accuracy.

Turning to pfgen-bench results, PLaMo 2.1 8B demonstrates performance comparable to PLaMo-100B, clearly showcasing its strong Japanese language generation capabilities. It also outperforms models of comparable size.\\

\noindent\textbf{Japanese-specific Knowledge: JMMLU Japanese-language Questions}\
JMMLU includes a special set of Japanese-language questions called ``Japanese-specific tasks'' that are not translations from MMLU.
These four task categories—idiomatic expressions, civics, Japanese geography, and Japanese history—collectively test knowledge that is less commonly assessed in English.
These performance results on JMMLU's Japanese-specific questions allow us to evaluate each model's proficiency in Japanese-specific knowledge.

\begin{table}[t]
    \centering
    \caption{Results of JMMLU's Japanese-specific Tasks}
    \begin{tabular}{lrrrrr}
        \toprule
        Model &  Average & \makecell{Japanese\\Idiom} & \makecell{Japanese\\Civics} & \makecell{Japanese\\Geography} & \makecell{Japanese\\History} \\
        \midrule
        Qwen/Qwen2.5-7B & 0.77 & 0.92 & 0.85 & 	0.77 & 	0.52 \\
        Qwen/Qwen3-8B-Base & 0.79 & 0.93 & 0.85 & 	0.86 & 0.52 \\
        tokyotech-llm/Llama-3.1-Swallow-8B-v0.2 & 0.87 & 0.97 & 0.89 & 0.91 & 0.71 \\
        sbintuitions/sarashina2-7b & 0.67 & 0.73 & 0.73	& 0.54 & 0.66 \\
        PLaMo 2 8B & 0.85 & 0.97 & 0.87 & 0.85 & 0.71 \\
        PLaMo 2.1 8B & 0.87 & 0.95 & 0.89 & 0.91 & 0.75 \\
        \midrule
        Qwen/Qwen2.5-32B & 0.84 & 0.96 & 0.92 & 0.87 & 0.62 \\
        abeja/ABEJA-Qwen2.5-32b-Japanese-v0.1  & 0.91 & 0.97 & 0.94 & 0.96 & 0.77 \\
        PLaMo 2 31B  & 0.90 & 0.97 & 0.92 & 0.93 & 0.78 \\
        \midrule
        pfnet/plamo-100b  & 0.85 & 0.94 & 0.87 & 0.86 & 0.74 \\
        \bottomrule
    \end{tabular}
\end{table}

PLaMo 2.1 8B demonstrates performance nearly on par with or exceeding PLaMo-100B on Japanese-specific tasks within JMMLU.
While PLaMo 2 uses English-translated data for its training dataset, these results confirm that this approach does not result in the loss or degradation of Japanese-specific knowledge.

When comparing models of the same size, PLaMo 2.1 8B outperforms both Qwen/Qwen3-8B-Base and sarashina2-7b, while achieving comparable accuracy to Llama-3.1-Swallow-8B.
The performance gap between Qwen/Qwen3-8B-Base and PLaMo 2.1 8B can be attributed to differences in Japanese-specific training data. The reason for the difference between PLaMo 2.1 8B and sarashina2-7b remains unclear, though the overall performance disparity across JMMLU may manifest in Japanese-specific tasks as well.

\subsubsection{The Effects of Pruning}

We will examine the effects of pruning and its benefits by comparing models such as PLaMo 2 8B and PLaMo 2.1 8B.
\Cref{tab:resouce-performance} summarizes the computational resources (FLOPs) used for training and the corresponding performance on JMMLU benchmarks. For reference, we also include performance data for a 31B model. Note that the training computational requirements for PLaMo 2.1 8B do not include those of the original PLaMo 2 31B model.

\begin{table}[t]
    \centering
    \caption{Computing Resource vs Model Performance. For PLaMo 2.1 8B, the computing resource excludes those of the original PLaMo 2 31B model used for pruning, but includes both the training computing resource for the 8B model itself and the inference computing resource for the 31B model used during distillation.}
    \label{tab:resouce-performance}
    \begin{tabular}{lrrr}
        \toprule
        Model   & Trainig Tokens & Computing Resource ($10^{18}$ FLOPs) & JMMLU (5 shot, acc)\\
        \midrule
        PLaMo 2 8B & 6T & 288,000 & 0.572 \\
        PLaMo 2.1 8B & 500B & 55,000  & 0.672 \\
        PLaMo 2 31B & 2T & 	372,000 & 0.635 \\
        \bottomrule
    \end{tabular}
\end{table}

PLaMo 2.1 8B demonstrates superior performance among the three models despite requiring significantly less training compute compared to both PLaMo 2 8B with and without pruning.
This suggests that pruning and re-training represent highly effective approaches from the perspective of required training compute. Note that pruning requires an original model, and the corresponding training process uses more computational resources than the PLaMo 2 8B baseline. However, high-performance large models are inherently necessary regardless of whether pruning is applied, and we believe it is reasonable to consider this additional compute resource separately from the primary training requirements.

This remarkable pruning efficiency is particularly significant when reducing model size. It is well established that, for a given model size, further increasing the number of training tokens eventually yields diminishing returns in performance improvement \citep{gadre2024languagemodelsscalereliably}. The ability to achieve high performance with significantly fewer training tokens through pruning and re-training effectively mitigates this issue, potentially enabling the training of high-performance models without increasing model size.

Furthermore, we believe that pruning offers benefits when considering the entire pre-training process.
Depending on the requirements of GPUs and other hardware, the optimal LLM size varies. For instance, when prioritizing maximum performance, one would naturally opt for a massive model, while for edge device deployment, smaller models are typically more suitable.

Consequently, LLM pre-training requires creating multiple model variants. Traditionally, meeting this requirement necessitated running multiple large-scale training experiments independently. Running numerous similar large-scale experiments significantly increases the overall computational resources required for pre-training.
Through the pruning method introduced here, we have achieved an efficient and waste-free approach to preparing high-performance pre-training model variants.

\section{Post-training}
\label{sec:post_training}

In the PLaMo 2 series' post-training, we introduced several enhancements to improve model performance to the post-training pipeline of PLaMo-100B-Instruct~\citep{plamo100b}, which consists of supervised fine-tuning (SFT) and direct preference optimization (DPO)~\citep{dpo}.

In the following sections, we describe the key enhancements we introduced:
continuous pre-training for better handling of long-context tasks (\Cref{subsec:cpt}), generation of high-quality Japanese synthetic datasets using LLMs (\Cref{subsec:posttrain_datasets}), and post-training pipeline refinements such as post-SFT model merging and algorithmic extensions of DPO (\Cref{subsec:posttrain_enhancements}).
In \Cref{subsec:posttrain_evaluation}, we present the results of several LLM evaluation benchmarks, which demonstrate the competitive performance of our post-trained models, PLaMo 2.1-8B and PLaMo 2.0-31B, on Japanese language tasks.

\subsection{Continual Pre-training for Long-context Support}
\label{subsec:cpt}

This section describes the continual pre-training (CPT) we performed after the initial pre-training of PLaMo 2 to enable better handling of longer contexts.
We first identify the architectural limitations of the original design of PLaMo 2 in the long-context retrieval task, then explain the architectural modifications and training methods we adopted to address those issues, along with our evaluation process and final model selection.

\subsubsection{Limitations of Retrieval Capabilities in Samba Architecture}
\label{subsubsec:limitations_arch}
As described in \Cref{subsec:model_arch}, PLaMo 2 employs a hybrid model architecture inspired by Samba designed with an emphasis on computational efficiency and scalability. 
This architecture alternates between the Selective State Space Model (SSM) and sliding window attention (SWA).
The design philosophy combines the advantages of SSM, including linear-time computational complexity, efficient sequential processing, and stable sequence-length extrapolation capabilities, with the ability of SWA to capture complex local non-Markovian dependencies~\citep{samba}.
This approach aimed to mitigate the challenges faced by pure Transformer models, such as quadratic growth in computational requirements with input length and significant memory consumption due to expanding key-value caches during inference.

\begin{figure}[h]
    \centering
    \begin{minipage}[b]{.47\hsize}
        \centering
        \includegraphics[width=.8\linewidth]{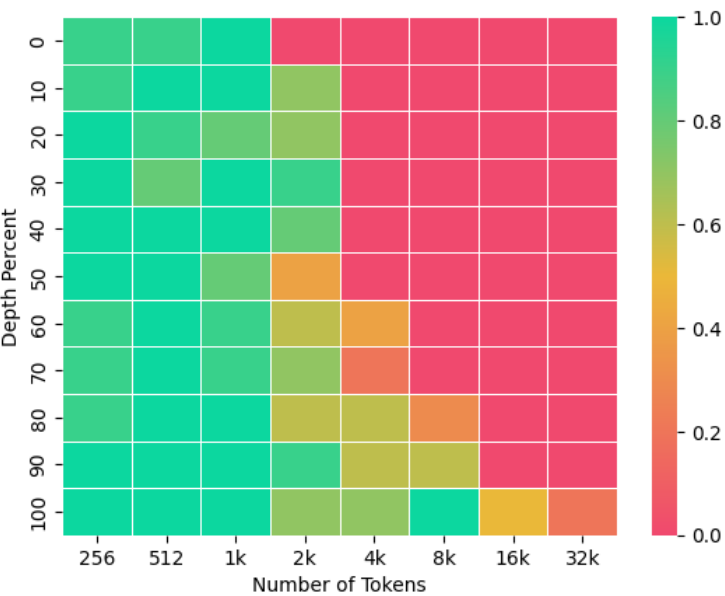}
        \subcaption{
            Phonebook task accuracy.
        }
        \label{fig:plamo1b_phonebook}
    \end{minipage}
    \;\;
    \begin{minipage}[b]{.47\hsize}
        \centering
        \includegraphics[width=.8\linewidth]{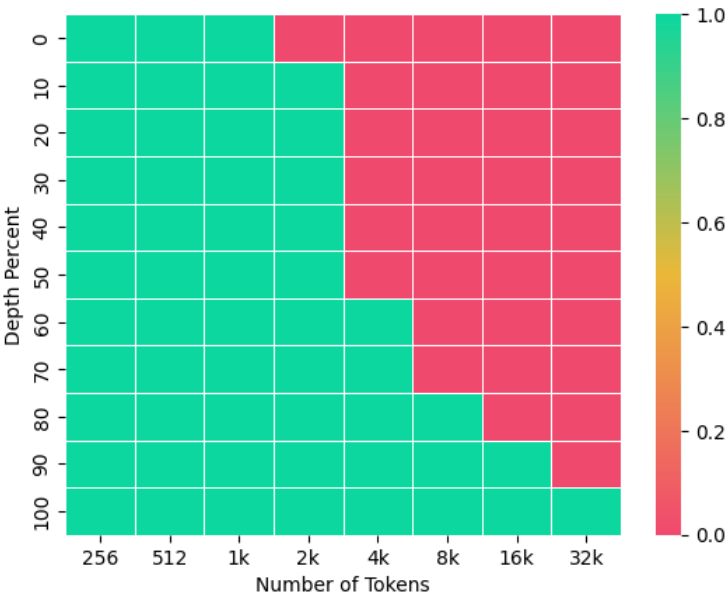}
        \subcaption{
            Passkey Retrieval task accuracy.
        }
        \label{fig:plamo1b_passkey}
    \end{minipage}
    \caption{
        Long-context performance for PLaMo 2-1B.
    }
    \label{fig:long_context_performance_plamo1b}
\end{figure}

However, initial post-training evaluations of PLaMo 2 revealed fundamental architectural constraints.
In particular, we observed performance issues in the tasks like needle-in-a-haystack, specifically, the ability to accurately retrieve and extract specific information from long contexts. 
Specifically, in both the Phonebook~\citep{phonebook} and the Passkey Retrieval~\citep{passkey_retrieval} tasks, the model consistently failed these tasks when the target information was located beyond the SWA window size of $2\,048$ tokens distanced from the output tokens as shown in \Cref{fig:long_context_performance_plamo1b}, where we used our smallest 1B model. 
This implies that the fixed window size of SWA creates a barrier to information access.

\begin{figure}[h]
    \centering
    \begin{minipage}[b]{.47\hsize}
        \centering
        \includegraphics[width=.8\linewidth]{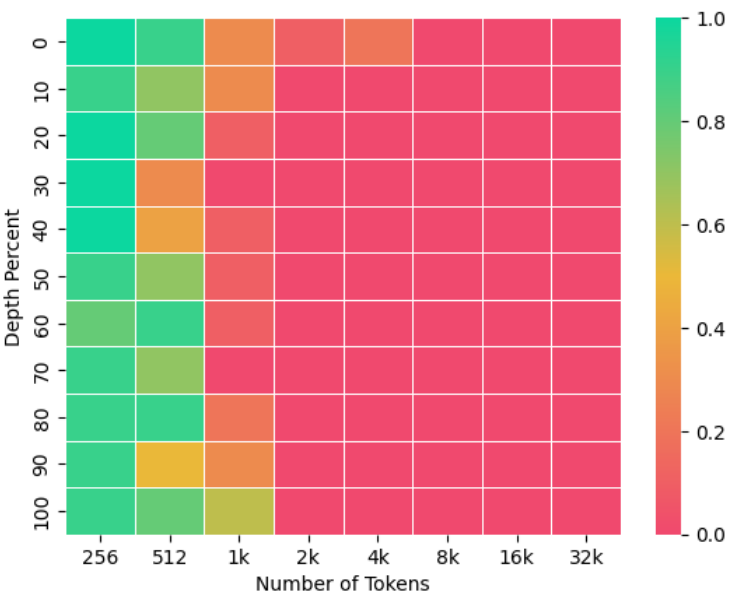}
        \subcaption{
            Phonebook task accuracy.
        }
        \label{fig:falcon3_phonebook}
    \end{minipage}
    \;\;
    \begin{minipage}[b]{.47\hsize}
        \centering
        \includegraphics[width=.8\linewidth]{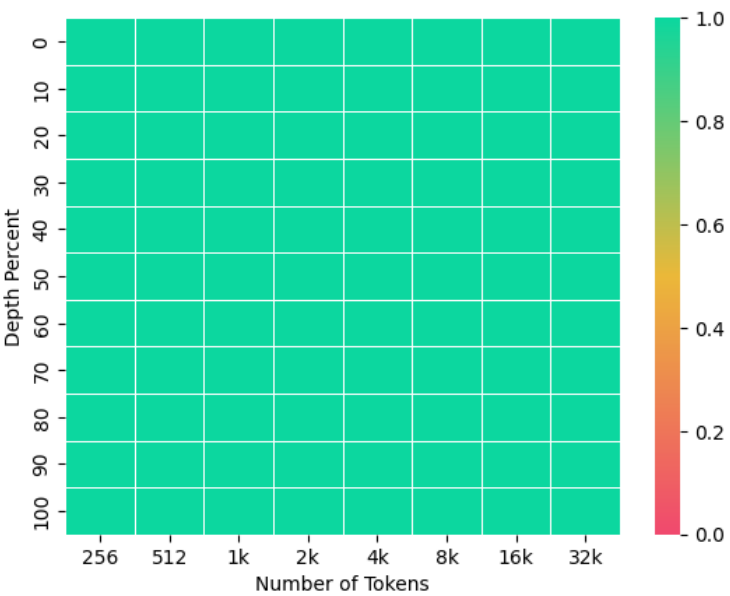}
        \subcaption{
            Passkey Retrieval task accuracy.
        }
        \label{fig:falcon3_passkey}
    \end{minipage}
    \caption{
        Long-context performance for Falcon3-Mamba-7B-Instruct.
    }
    \label{fig:long_context_performance_falcon3}
\end{figure}

This challenge appears to be not unique to PLaMo 2, but rather stems from common characteristics of both the SSM and SWA architectures. 
SSMs are known by their nature to have difficulty maintaining lossless retention of specific information over time, as their architecture compresses past information into a fixed-length hidden state~\citep{phonebook}.
Their strength lies in information summarization and compression rather than high-resolution recall of information at any arbitrary position~\citep{hymba}.
This observation was confirmed through our own independent evaluation of a pure SSM, Falcon3-Mamba model~\citep{Falcon3}.
While Falcon3-Mamba demonstrated perfect performance on the Passkey Retrieval task, it showed poor performance on the Phonebook task even with relatively short context lengths of around 1K tokens as shown in \Cref{fig:long_context_performance_falcon3}, suggesting that SSMs fundamentally struggle with this type of task. 
This observation aligns with claims made by some studies suggesting that models using pure SSM or SWA alone perform worse than full attention for needle-in-a-haystack-type tasks (long-context information retrieval)~\citep{retrieval_head}.

Based on these analyses, we concluded that the difficulty of the Phonebook task may not be due to insufficient training or tuning issues, but rather stems from constraints inherent to the architectural choice, namely, the combination of SSM and SWA which localizes the attention field in the attention layer.
Consequently, we determined that rather than continuing the training with this architecture as is, a more substantial approach involving architectural modifications itself was necessary to overcome this limitation in retrieval capability.

\subsubsection{Implementation of Full Attention Mechanism}
\label{subsubsec:impl_full_attn}
To overcome the limitations in retrieval capability mentioned earlier, we designed a dedicated CPT phase to support long-context inputs, incorporating targeted architectural modifications. 
Specifically, we expanded the window size of the SWA layers in PLaMo 2 to match the desired context length before starting CPT. 
Specifically, we decided to enlarge the window size of the SWA layers in PLaMo 2 to match the desired context length before starting CPT. This decision deliberately sacrifices the linear-time computational efficiency provided by SWA during this specific training phase, instead prioritizing the model's ability to access the entire context unrestricted--essentially granting it high-resolution recall capabilities.

\subsubsection{Details of Continual Pre-training}
\label{subsubsec:details_cpt}
The CPT for long-context capabilities was fundamentally designed to improve next token prediction accuracy, maintaining the same problem formulation as conventional pre-training. However, beyond architectural modifications, there were also differences in both dataset selection and training configurations, which we summarize below.

\paragraph{Dataset}
For the CPT, we used a subset of the corpus created for pre-training. 
This approach was designed to preserve the consistency of the data distribution while minimizing the risk of catastrophic forgetting of general knowledge.

\paragraph{RoPE Frequency}
As the core technology to expand the context length from 2K to 32K, we adopted Adjustable Base Frequency RoPE~\citep{abf_rope}.
For our 32K context length requirement, we set the RoPE base frequency parameter, which popular packages call \texttt{rope\_theta}, to $1\,000\,000$, matching the configuration used in Gemma-3~\citep{gemma3}.

\subsubsection{Empirical Evaluation and Checkpoint Selection}
\label{subsubsec:cpt_eval}
The effectiveness of the CPT was verified through empirical evaluation, and the final model selection was conducted while considering the balance among multiple performance metrics.

\begin{figure}[h]
    \centering
    \includegraphics[width=.5\linewidth]{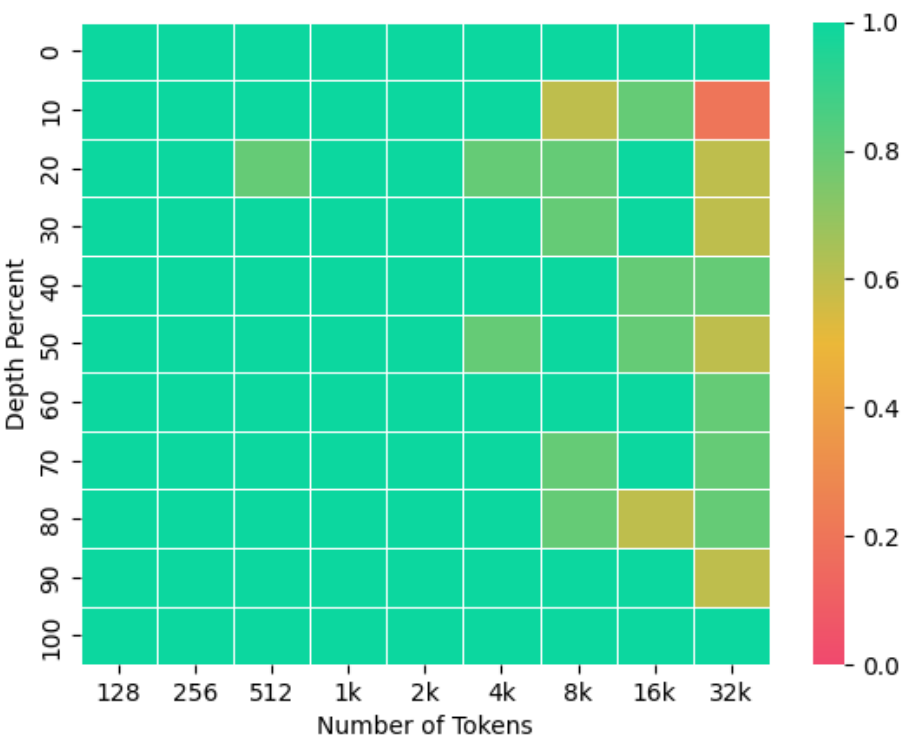}
    \caption{
        Long-context performance (Phonebook task accuracy) for PLaMo 2-31B after continual pre-training with 1.75B tokens.
    }
    \label{fig:long_context_performance_plamo30b}
\end{figure}

\paragraph{Performance Improvement on Target Task}
The direct effect of resizing the SWA window size before performing CPT became clearly evident in the performance of the Phonebook task.
With any model size, significant performance improvements on the Phonebook task began appearing after approximately 1 billion tokens. 
The accuracy on the Phonebook task immediately after CPT for PLaMo 2-31B is shown in Figure~\ref{fig:long_context_performance_plamo30b}.

\paragraph{Checkpoint Selection}
Our final model selection process did not simply involve choosing the checkpoint with the highest Phonebook benchmark score.
Even if we succeed in enhancing the ability to extract specific information from long sequences, it would be undesirable if this comes at the cost of the model's general language capabilities.
Therefore during CPT, we saved checkpoints every 0.25B tokens and calculated both Phonebook task scores and scores for pfgen-bench--a benchmark specifically designed to assess more generalized Japanese text generation capability--for all these checkpoints.
We then selected the checkpoint that achieved the optimal balance between these two performance metrics.

\subsubsection{Impact on General Long-context Tasks}
\label{subsubsec:cpt_impact_on_general_tasks}
The acquisition of long-context capability through this CPT not only improved the performance on the Phonebook task but also contributed to enhance the performance of other long-context comprehension tasks.
For instance, the 2B model showed significant improvement in LongBench \citep{Bai2024ACL} scores through both the CPT and subsequent SFT, surpassing competing models like Qwen2.5-1.5B \citep{qwen25technicalreport}.
These results suggest that the high-resolution recall capability acquired during this CPT demonstrates some generalization potential for diverse long-context understanding tasks.

\subsection{Post-training Datasets}
\label{subsec:posttrain_datasets}

There are less publicly available high-quality Japanese datasets for LLM's post-training compared with English.
To address this issue, PLaMo 2 made extensive use of LLM-generated datasets.
For example, we generated a dataset to enhance the instruction-following capabilities.
We first generated instruction-guided questions and responses using LLMs by combining random questions with various response format instructions.
We then filtered the data through both programmatic verification of instruction following and LLM-as-a-Judge assessment of response quality, producing a dataset used in both SFT and DPO.

We also synthesize and re-generate SFT and DPO datasets from our previous attempt \citep{plamo100b} to enhance PLaMo's language-skills. 
In the following section, we share that we generated LLM's response texts based on two publicly available datasets to diversify our post-training datasets.

\subsubsection{llm-jp-instructions}
As a single-turn user-assistant conversational dataset created by native Japanese speakers, we tried to use the original dataset.\footnote{\url{https://huggingface.co/datasets/llm-jp/llm-jp-instructions}}
However, we found that the original human-annotated responses had a few issues, such as markdown syntax error.
To mitigate such issues, we generated responses using an LLM distributed under a permissive license.
After that, we manually removed prompt-response pairs that had obvious problems, e.g., containing a nonexistent URL.

\subsubsection{Wildchat-1M}
As a multi-turn conversation dataset, we used the user prompts of Wildchat-1M \citep{Zhao2024ICLR}.
Note that we removed assistant texts from the original dataset generated by OpenAI's ChatGPT to avoid violating the Terms of Use.
As a pre-processing step, we retained only conversations in Japanese or English, whose language was specified in the original dataset since our LLMs mainly focus on Japanese and English languages.
In addition, we performed fuzzy deduplication using a similar procedure performed by \citet{Shen2023arXiv}. 
Concretely, we constructed a similarity graph, where each node represents one example: user-text.
The edge exists if two nodes have a similarity of $0.8$ or higher.
For each connected component, sample one example randomly and discard the others.
To compute similarity, we computed the minhash signature at the example level using $128$ permutation functions \citep{Broder2000CPM}\footnote{We used \url{https://ekzhu.com/datasketch/index.html}.} and a 13-gram character feature using \texttt{nltk} \citep{bird2004ACL} after text normalization.
For faster similarity computation, we used \texttt{faiss} \citep{douze2024faiss}.
Given user prompts, we generated assistant text using an LLM distributed under a permissive license.

To increase the number of Japanese conversations, we translate generated English conversations into Japanese using our in-house LLM-based translation model, which is a prototype of the PLaMo 2-Translation model.\footnote{\url{https://huggingface.co/pfnet/plamo-2-translate}}
Concretely, we generated $16$ translated responses and selected the best one that achieved the highest MetricX-24 \citep{Juraska2024WMT} score.
We used these datasets as a part of SFT dataset.

Using the generated data, we generated preference data for DPO using our intermediate instructed model using our 31B model.
To do so, we filtered out too long examples for our reward model, Nemotron4-340B-reward \citep{Nvidia2024arXiv}.
We generated six responses to conversation texts whose last LLM assistant was discarded.
We selected the highest-reward text as the chosen response and the lowest-reward text as the rejected response in the seven responses.
Recall that one of them was generated for SFT, and the others six were generated using the intermediate PLaMo 2-31B.
We found that some chosen and rejected reward scores were too close; thus, we removed preference pairs from the dataset if the reward gap is too small, inspired by \citet{Amini2024FindingsACL}.

To further improve our post-trained model, we replaced the low-quality translated responses in the SFT dataset with the DPO's chosen responses.
As empirically shown by \citet{Tang2024arXiv}, the on-policy approach is more effective than the off-policy approach.
This replacement imitated such an on-policy approach and improved the benchmark scores in our preliminary experiments.

\subsection{Post-training Procedures}
\label{subsec:posttrain_enhancements}

This section details the methodological improvements implemented for post-training PLaMo 2 beyond dataset modifications.

\subsubsection{Chat Template Design}

A chat template is used to feed user-assistant conversations into an LLM. To handle multi-turn dialogues, multiple string inputs are encoded into a single token sequence, requiring decisions regarding speaker representation, prefixes, separators, and other formatting elements.

PLaMo 1.0 employed a Markdown-like format called Alpaca \citep{alpaca}, with its separator specifically set to \verb|"\n\n###"| (equivalent to Markdown's h3 header). We implemented a dedicated token for the separator for PLaMo 2.0, referring to ChatML \citep{openai-chatml} where not one but two special tokens are used. During pre-training, PLaMo 2.0 was trained on a diverse dataset encoded using this special token \verb/<|plamo:op|>/, including question-answering data. This pre-training phase already incorporated certain chat capabilities, enabling seamless transition to post-training using chat templates with this token. This approach is expected to reduce potential confusion for the LLM when processing Markdown-style inputs in addition to minor optimization in token count.

Additionally, we enhanced compatibility with tasks like the Jaster benchmark that clearly distinguish between instructions and inputs. When interpreting this as a conversation, two approaches were considered: treating inputs as user inputs and instructions as higher-level role inputs (system prompts), or treating instructions as user inputs while considering task inputs as contextual information. We finally adopted the latter approach. Although \texttt{input} is non-standard for a role name in chat APIs, Alpaca implemented it by \verb|### Input:| and we maintained consistency with PLaMo 2.0's pre-training. For example, during pre-training, data transformation tasks were tagged with \texttt{input} and \texttt{output} roles, which now aligns exactly with our chat representation between \texttt{input} and \texttt{output} roles.

\subsubsection{Model Merge at the SFT Stage}

For an efficient SFT pipeline, we merged different supervised fine-tuned parameters from two partially shared datasets \citep{Li2022arXiv,Cohere2025arXiv} rather than performing a single SFT on the union of datasets.
One dataset was a general SFT dataset, and the other was a publicly available dataset dominating math and coding \citep{Bercovich2025arXiv}, where we removed examples generated by using non-permissive models: Llama 3 families and Qwen2.5-72B.
In addition, we removed examples that had a reasoning thought because our model does not focus reasoning capability.
Note that we added about 10\% of the general dataset into the other dataset by following \citet{Cohere2025arXiv}. 
We merged these two parameters with a weighted mean with a single global proportion \citep{Wortsman2022ICML}.

\subsubsection{DPO Enhancements}

We further enhanced the post-SFT merged model through DPO training.
During this process, we incorporated two extensions to the DPO loss function: a regularization mechanism to suppress excessively long outputs~\citep{park2024disentangling} and the addition of SFT loss for chosen responses~\citep{pang2024iterative}, both of which proved effective in preventing performance degradation on specific tasks.
Following DPO, we also performed model merging across multiple model configurations trained under different settings.

\subsection{Evaluation}
\label{subsec:posttrain_evaluation}

We conducted comprehensive performance evaluations of both PLaMo 2.1-8B and PLaMo 2.0-31B after post-training. 
We used Jaster, M-IFEval Japanese, Elyza-Tasks-100, and pfgen-bench as benchmarks. For comparison of the PLaMo 2.1-8B model, we evaluated Qwen2.5-7B-Instruct, Qwen3-8B, and Llama-3.1-8B-Instruct as open-weight models within the same parameter range. For comparison of PLaMo 2.0-31B, we evaluated the preceding PLaMo version PLaMo 1.0 Prime (PLaMo-100B~\citep{plamo100b}), around 30B-class open-weight models (Qwen2.5-32B-Instruct, Mistral-Small-3.1-24B-Instruct-2503, and gemma-3-27b-it), and OpenAI's gpt-4o-mini.

\subsubsection{Jaster}

Jaster~\citep{Han_llm-jp-eval_2024} is a benchmark for evaluating the performance of Japanese using a variety of datasets. Basically, it consists of short answer tests evaluating the model's knowledge about Japan and the Japanese language.

For this evaluation, we used version 1.4.1\footnote{\url{https://github.com/llm-jp/llm-jp-eval/tree/v1.4.1}} with the temperature set to 0, and the generation was performed in completion mode. Since gpt-4o-mini did not provide a completions API, we generated responses using the chat completions API with prompts as single-turn chats. The average scores across 4-shot conditions (AVG) and category-specific scores are detailed in \Cref{jaster-8b,jaster}.

\begin{table}[t]
    \centering
    \caption{Jaster benchmark scores for 8B-class models.}
    \label{jaster-8b}
    \scriptsize
    \begin{tabular}{lrrrrrrrrrr}
    \toprule
    Model Name & AVG & EL & FA & HE & MC & MR & MT & NLI & QA & RC \\
    \midrule
    PLaMo 2.1-8B & \textbf{0.626} & 0.329 & \textbf{0.520} & 0.632 & \textbf{0.855} & 0.752 & \textbf{0.881} & \textbf{0.852} & 0.544 & \textbf{0.890} \\
    Qwen2.5-7B-Instruct & 0.501 & 0.448 & 0.195 & 0.390 & 0.660 & 0.670 & 0.742 & 0.728 & 0.402 & 0.770 \\
    Qwen3-8B & 0.593 & \textbf{0.542} & 0.250 & \textbf{0.660} & 0.787 & \textbf{0.810} & 0.853 & 0.742 & 0.410 & 0.876 \\
    Llama-3.1-8B-Instruct & 0.547 & 0.457 & 0.236 & 0.570 & 0.757 & 0.760 & 0.751 & 0.628 & \textbf{0.440} & 0.872 \\
    \bottomrule
    \end{tabular}
\end{table}

\begin{table}[t]
\centering
\caption{Jaster benchmark scores for 31B-class models. Note that PLaMo 1.0 Prime has 100B parameters.}
\label{jaster}
\scriptsize
\begin{tabular}{lrrrrrrrrrr}
\toprule
Model Name & AVG & EL & FA & HE & MC & MR & MT & NLI & QA & RC \\
\midrule
PLaMo 2.0-31B & \textbf{0.665} & 0.466 & \textbf{0.524} & 0.645 & 0.860 & 0.840 & \textbf{0.884} & \textbf{0.852} & \textbf{0.662} & \textbf{0.915} \\
PLaMo 1.0 Prime & 0.620 & 0.381 & 0.538 & 0.545 & \textbf{0.890} & 0.760 & 0.879 & 0.808 & 0.500 & 0.895 \\
Qwen2.5-32B-Instruct & 0.659 & \textbf{0.609} & 0.285 & \textbf{0.795} & 0.880 & 0.960 & 0.865 & 0.792 & 0.526 & 0.876 \\
Qwen3-32B & 0.567 & 0.455 & 0.192 & 0.740 & 0.880 & 0.820 & 0.585 & 0.768 & 0.362 & 0.873 \\
Mistral-Small-3.1-24B-Instruct-2503 & 0.649 & 0.540 & 0.331 & 0.755 & 0.853 & 0.920 & 0.870 & 0.770 & 0.608 & 0.846 \\
gemma-3-27b-it & 0.579 & 0.573 & 0.335 & 0.370 & 0.727 & \textbf{0.970} & 0.873 & 0.566 & 0.572 & 0.804 \\
gpt-4o-mini & 0.635 & 0.601 & 0.333 & 0.660 & 0.880 & 0.940 & 0.871 & 0.672 & 0.508 & 0.881 \\
\bottomrule
\end{tabular}
\end{table}

In both model sizes, the PLaMo 2 series achieved the highest AVG scores across all categories, outperforming other models in most evaluation metrics. This confirms that the PLaMo 2 models, which were specifically trained on extensive Japanese data, have acquired superior Japanese language capabilities.

Regarding the MR (Mathematical Reasoning) category, which consists of basic arithmetic tasks that posed challenges for PLaMo 1.0 Prime, PLaMo 2.0-31B only achieved scores that were about 0.08 to 0.13 lower than other models. Although significant improvements have been made from PLaMo 1.0 Prime through enhancements in pre-training and other areas, we intend to continue making further improvements in the future.

\subsubsection{M-IFEval Japanese Subset}

M-IFEval~\citep{Dussolle2025MIFEval} is a multilingual version of the IFEval benchmark~\citep{ifeval}, which evaluates an instruction-following ability of LLMs. We used the Japanese subset\footnote{\url{https://github.com/lightblue-tech/M-IFEval/blob/main/data/ja_input_data.jsonl}} of this benchmark. Like IFEval, this benchmark consists of relatively simple instructions that can be automatically verified, and the Japanese subset includes Japanese-specific examples such as:

\begin{itemize}
    \item Character count instructions: ``600文字以上で答えてください" (Please provide your response in 600 or more characters.)
    \item Character type specifications: ``ひらがなだけを使って答えてください" (Answer using only hiragana characters.)
    \item Kanji usage instructions: ``40字未満に漢字の使用回数を抑えて解説してください" (Limit your use of kanji to fewer than 40 characters in your explanation.)
    \item Punctuation instructions: ``応答全体で句点「。」を用いずに、番号付きリストの形で説明してください" (Describe your response in numbered list format without using any periods ``。".)
\end{itemize}

We set the generation temperature to 0, following the setting in lm-evaluation-harness\footnote{\url{https://github.com/EleutherAI/lm-evaluation-harness/blob/v0.4.9/lm_eval/tasks/ifeval/ifeval.yaml\#L12}}, and performed generation using chat completions. Since the greedy decoding (temperature = 0) is not recommended for Qwen 3's thinking mode\footnote{\url{https://huggingface.co/Qwen/Qwen3-32B\#best-practices}}, we use the average results from five runs with the recommended sampling parameters. The evaluation results are presented in \Cref{ifeval-ja-8b,ifeval-ja}.

\begin{table}[t]
    \centering
    \caption{M-IFEval Japanese subset benchmark scores for 8B-class models.}
    \label{ifeval-ja-8b}
    \scriptsize
    \begin{tabular}{lccc}
        \toprule
        Model Name & AVG & strict prompt level & strict instruction level \\
        \midrule
        PLaMo 2.1-8B & \textbf{0.630} & \textbf{0.600} & \textbf{0.660} \\
        Qwen2.5-7B-Instruct & 0.463 & 0.430 & 0.496 \\
        Qwen3-8B (non-thinking) & 0.488 & 0.453 & 0.522 \\
        Qwen3-8B (thinking) & 0.559 & 0.530 & 0.588 \\
        Llama-3.1-8B-Instruct & 0.342 & 0.308 & 0.376 \\
        \bottomrule
    \end{tabular}
\end{table}

\begin{table}[t]
    \centering
    \caption{M-IFEval Japanese subset benchmark scores for 31B-class models. Note that PLaMo 1.0 Prime has 100B parameters.}
    \label{ifeval-ja}
    \scriptsize
    \begin{tabular}{lrrr}
        \toprule
        Model Name & AVG & Strict Prompt Level & Strict Instruction Level \\
        \midrule
        PLaMo 2.0-31B & \textbf{0.677} & \textbf{0.651} & \textbf{0.704} \\
        PLaMo 1.0 Prime & 0.342 & 0.308 & 0.376 \\
        Qwen2.5-32B-Instruct & 0.628 & 0.610 & 0.646 \\
        Qwen3-32B (non-thinking) & 0.576 & 0.541 & 0.611 \\
        Qwen3-32B (thinking) & 0.629 & 0.594 & 0.663 \\
        Mistral-Small-3.1-24B-Instruct-2503 & 0.497 & 0.459 & 0.535 \\
        gemma-3-27b-it & 0.574 & 0.541 & 0.606 \\
        gpt-4o-mini & 0.610 & 0.570 & 0.650 \\
        \bottomrule
    \end{tabular}
\end{table}

Focusing on \Cref{ifeval-ja}, PLaMo 1.0 Prime's scores were significantly lower than those of other models. This result aligns with feedback received after the PLaMo 1.0 Prime release, confirming that M-IFEval Japanese successfully measures one aspect of instruction-following performance in Japanese.

Both PLaMo 2.1-8B and PLaMo 2.0-31B show substantial improvements in instruction-following performance compared to PLaMo 1.0 Prime, achieving higher performance levels than other models in the same size category.
Other models fall behind in scores for Japanese-specific instructions (such as those regarding hiragana and kanji), demonstrating PLaMo 2.0-31B's strong Japanese language capabilities.
However, it should be noted that M-IFEval measures only simple instructions that can be automatically verified.
Evaluating and improving instruction-following performance for more complex instructions commonly encountered in real-world tasks remains an area for future research.

\subsubsection{ELYZA-tasks-100}

ELYZA-tasks-100~\citep{elyzatasks100} is a benchmark used to assess the ability to provide useful responses and follow instructions for complex tasks in Japanese.

\begin{figure}[t]
    \centering
    \includegraphics[width=0.8\textwidth]{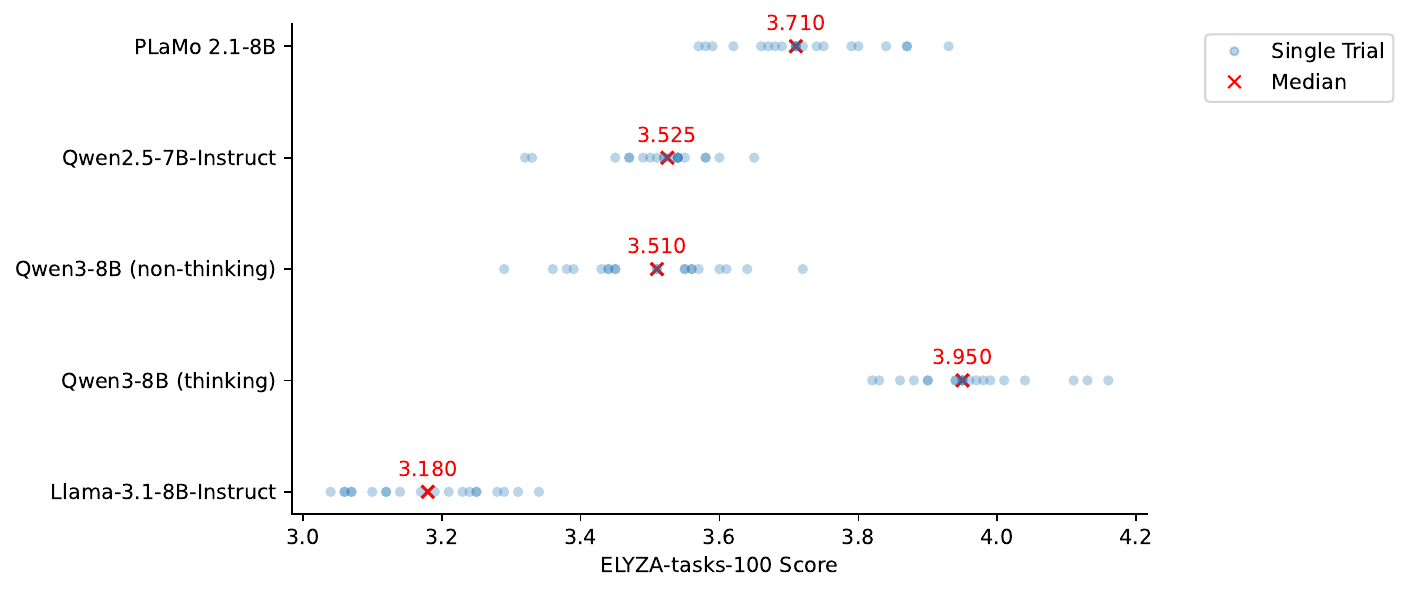}
    \caption{ELYZA-tasks-100 benchmark scores for 8B-class models.}
    \label{elyza-8b}
\end{figure}

\begin{figure}[t]
    \centering
    \includegraphics[width=0.8\textwidth]{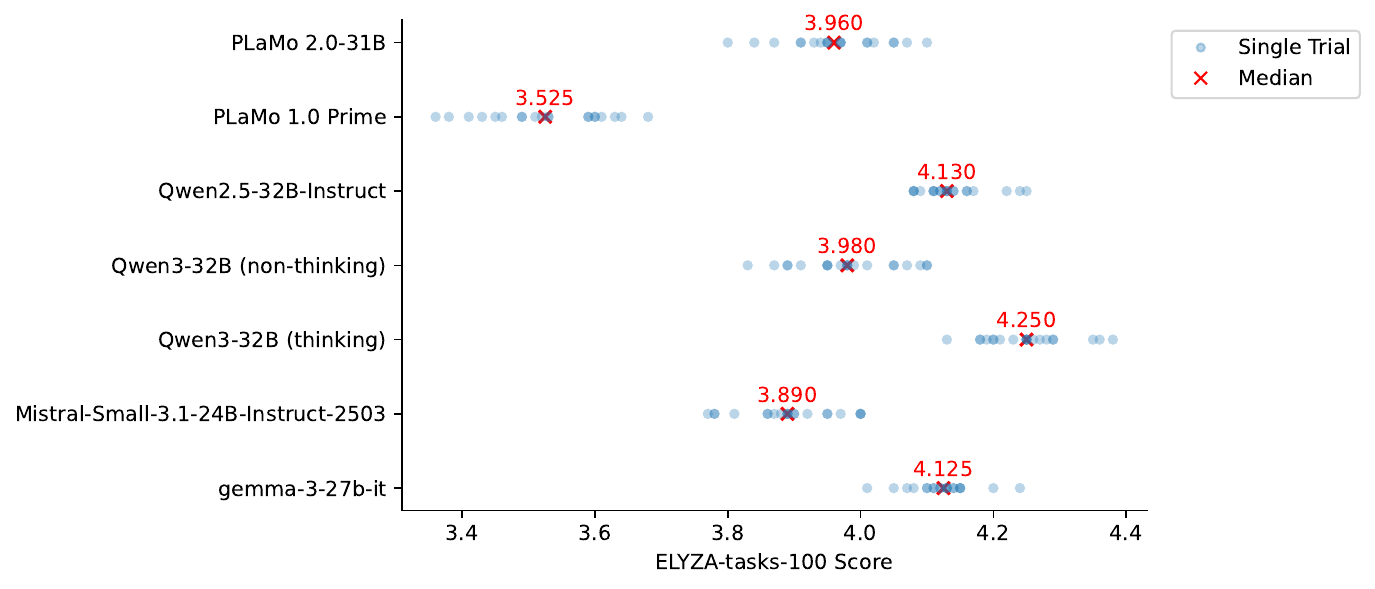}
    \caption{ELYZA-tasks-100 benchmark scores for 31B-class models. Note that PLaMo 1.0 Prime has 100B parameters.}
    \label{elyza}
\end{figure}

The scoring was conducted using automated evaluation with gpt-4o.
The prompts were based on the guidelines\footnote{\url{https://zenn.dev/elyza/articles/5e7d9373c32a98}} provided by ELYZA, the benchmark's developer.
To ensure consistent comparison of models with similar scores, we executed each evaluation 20 times with temperature set to 1.0 to account for evaluation variability (with the exception of the Qwen3 thinking mode, for which we used the official recommended sampling parameters).
The score distribution and median values are shown in \Cref{elyza-8b,elyza}.
Since gpt-4o was used for evaluation, gpt-4o-mini was excluded from consideration due to concerns about evaluation bias.

\Cref{elyza-8b} demonstrates that PLaMo 2.1-8B achieved particularly strong scores among the non-reasoning models evaluated in this comparison. \Cref{elyza} shows that PLaMo 2.0-31B showed a significant improvement in score compared to PLaMo 1.0 Prime.

Compared with Qwen3-32B (thinking mode), Qwen2.5-32B-Instruct, and gemma-3-27b-it, PLaMo 2.0-31B scored slightly lower. Upon reviewing the scores for each problem, it was observed that there were deductions due to incorrect answers in problems requiring multi-step reasoning.  This suggests that there is room for improvement in mathematical reasoning or other reasoning capabilities. We plan to address these areas through post-training improvements and expect further enhancement in future minor updates of PLaMo 2.

\subsubsection{pfgen-bench}

pfgen-bench~\citep{pfgen} is a Japanese text generation benchmark developed by us, with particular emphasis on evaluating fluency and accuracy in responses based on Japanese-specific common knowledge.
Based on preliminary experiments that showed significant score variations depending on generation parameters such as temperature and top\_p, we standardized the evaluation settings for this assessment: temperature = 0.0, top\_p = 0.98, num\_trials = 10, and mode = completion. However, since gpt-4o-mini did not provide a completions API, we evaluated it using mode = chat. 
The results are presented in \Cref{pfgen-8b,pfgen}.

\begin{table}[t]
    \centering
    \caption{pfgen-bench scores for 8B-class models.}
    \label{pfgen-8b}
    \scriptsize
    \begin{tabular}{lrrrrr}
        \toprule
        Model Name & Score & Length & Fluency & Truthfulness & Helpfulness \\
        \midrule
        PLaMo 2.1-8B & $\mathbf{0.893}$ ($\pm 0.016/\sqrt{10}$) & $103.1$ ($\pm 11.0$) & $\mathbf{0.964}$ & $\mathbf{0.960}$ & $\mathbf{0.755}$ \\
        Qwen2.5-7B-Instruct & $0.590$ ($\pm 0.024/\sqrt{10}$) & $100.1$ ($\pm 14.5$) & $0.685$ & $0.859$ & $0.225$ \\
        Qwen3-8B & $0.634$ ($\pm 0.009/\sqrt{10}$) & $96.6$ ($\pm 9.7$) & $0.753$ & $0.878$ & $0.271$ \\
        Llama-3.1-8B-Instruct & $0.526$ ($\pm 0.009/\sqrt{10}$) & $97.8$ ($\pm 35.2$) & $0.632$ & $0.844$ & $0.101$ \\
        \bottomrule
    \end{tabular}
\end{table}

\begin{table}[t]
    \centering
    \caption{pfgen-bench scores for 31B-class models. Note that PLaMo 1.0 Prime has 100B parameters.}
    \label{pfgen}
    \scriptsize
    \begin{tabular}{lrrrrr}
        \toprule
        Model Name & Score & Length & Fluency & Truthfulness & Helpfulness \\
        \midrule
        PLaMo 2.0-31B & $\mathbf{0.890}$ \,($\pm 0.012/\sqrt{10}$) & $104.4$ \,($\pm 11.3$) & $0.951$ & $0.952$ & $\mathbf{0.767}$ \\
        PLaMo 1.0 Prime & $0.846$ \,($\pm 0.015/\sqrt{10}$) & $107.4$ \,($\pm 19.4$) & $\mathbf{1.012}$ & $\mathbf{0.971}$ & $0.553$ \\
        Qwen2.5-32B-Instruct & $0.731$ \,($\pm 0.015/\sqrt{10}$) & $108.7$ \,($\pm 17.7$) & $0.840$ & $0.918$ & $0.435$ \\
        Qwen3-32B & $0.754$ \,($\pm 0.010/\sqrt{100}$) & $98.2$ \,($\pm 14.2$) & $0.840$ & $0.923$ & $0.497$ \\
        Mistral-Small-3.1-24B-Instruct-2503 & $0.795$ \,($\pm 0.013/\sqrt{10}$) & $103.6$ \,($\pm 12.5$) & $0.929$ & $0.952$ & $0.505$ \\
        gemma-3-27b-it & $0.786$ \,($\pm 0.009/\sqrt{10}$) & $98.6$ \,($\pm 14.3$) & $0.889$ & $0.932$ & $0.536$ \\
        gpt-4o-mini\* & $0.804$ \,($\pm 0.007/\sqrt{10}$) & $89.4$ \,($\pm 16.8$) & $0.874$ & $0.963$ & $0.574$ \\
        \bottomrule
    \end{tabular}
\end{table}

The results clearly show that PLaMo 2 achieves significantly higher scores than other models, demonstrating notable superiority in Japanese fluency and Japanese-specific common knowledge compared to models with similar parameter sizes.

\section{Inference Performance Optimization}

This section describes our efforts to enhance PLaMo 2's inference performance and to reduce computational costs through the use of an efficient inference framework and by employing quantization techniques.

First, we adopted the vLLM framework~\citep{kwon2023efficient} to enable efficient inference and further optimized the PLaMo 2 implementation by leveraging vLLM's capabilities. We also attempted model weight quantization using GPTQ~\citep{frantar2023optq} and KV cache quantization, and successfully improved inference performance while maintaining the accuracy levels reported in the previous section.

\subsection{vLLM Customization}
\subsubsection{PLaMo 2 Support in vLLM}
To enable the operation of PLaMo 2.0-31B within the vLLM framework, we first implemented the model architecture using the official vLLM Model API to ensure compatibility with hybrid architectures. vLLM supports both generic attention mechanisms and the Mamba kernel. However, since the Mamba kernel implementation is relatively new and Mamba-based architectures have not yet been as widely adopted in models as attention-based ones, we encountered several practical implementation challenges. For instance, the v0 engine is designed with the assumption of a transformer architecture, which means that the KV cache and Mamba state are managed through separate mechanisms, requiring careful handling during the model implementation.

Furthermore, PLaMo 2 incorporates subtle architectural differences from standard Mamba implementations. Specifically, the Mamba layers in PLaMo 2 differ from conventional implementations in two key aspects:
\begin{enumerate}
    \item a linear projection operation is applied to the state parameters between the causal conv1d and selective scan operators, and
    \item there is no RMSNorm layer included after the selective scan operator.
\end{enumerate}
These differences necessitated the development of a custom Mamba layer. Additionally, while developing the PLaMo 2 implementation, we identified and addressed several issues related to memory leaks and improper allocation of the Mamba state within vLLM.

\subsubsection{Model Parallel}
In our model implementation, we incorporated support for model parallelism techniques including pipeline parallelism~\citep{harlap2018pipedream} and tensor parallelism~\citep{shoeybi2019megatron}. Pipeline parallelism is relatively straightforward to implement—by constructing models using vLLM's make\_layers interface, each layer can be naturally distributed across multiple GPUs. However, tensor parallelism represents a more complex approach. In general, it involves replacing standard Linear layers with parallelization-optimized alternatives based on the specific use case requirements.

\subsubsection{Chunked Prefill}
To further optimize runtime and hardware utilization, we adopted the chunked prefill technique. In typical LLM inference, the prefill phase is compute-bound, while the decode phase is memory bandwidth-bound. By interleaving prefill and decode requests within the same batch, chunked prefill facilitates a more balanced workload and enhances overall hardware utilization. However, we observed that this interleaving negatively impacted the performance of the Mamba layer. To address this issue, we explicitly modified the Mamba layer to separately process prefill and decode sequences, thereby restoring and enabling the performance gains achieved through chunked prefill. As a result, we observed performance improvements across Ampere, Ada, and Hopper GPUs.

\subsubsection{torch.compile Integration}
To enhance inference efficiency for PLaMo 2.0-31B, we selectively applied torch.compile to the model's attention components using the TorchInductor backend. In the vLLM environment, torch.compile primarily boosts performance by fusing individual PyTorch operations into fewer, more efficient Triton kernels. This process reduces memory traffic, enhances GPU occupancy, and minimizes kernel launch overhead.

Due to the current incompatibility of Mamba blocks with torch.compile, we employed piecewise compilation to focus solely on the attention layers. Within each attention block, we enabled combo\_kernels to horizontally fuse independent operations, such as Q/K normalization, into a single kernel. This aggressive fusion strategy significantly reduces the total kernel count and improves overall execution efficiency.

\subsection{Quantization}
This chapter evaluates the performance of quantization techniques applied to the PLaMo 2.0-31B model architecture, focusing on maintaining a cost-effective yet stable inference platform. To fully utilize the 32k context capacity of PLaMo 2.0-31B, substantial computational resources and memory capacity are required for inference. Specifically, running inference in 16-bit floating point format requires 63GB for model weights and 110KB for the KV cache per token, totaling 3.6GB for the 32k context. Additionally, extra memory is needed for activations. This significantly exceeds the memory capacity of widely used GPUs such as L40S and A40. Furthermore, when accounting for memory usage by the KV cache and activation functions, even GPUs with 80GB capacity face significant challenges in performing efficient inference on a single card.

To address this challenge, we implemented quantization of both model weights and KV cache to reduce computational costs, then evaluated the resulting performance.

\subsubsection{Weight Quantization}

To compress the model weight size, we applied INT4 quantization using GPTQ~\citep{frantar2023optq}. To ensure parallel efficiency, we applied quantization to all parameters except for the conv1d layers and a few very small linear layers. This reduced the model size by 70\% to 17GB, significantly facilitating large-scale model inference on general-purpose hardware.

As calibration datasets, we utilized the C4 corpus~\citep{raffel2020exploring} and ToolACE~\citep{liu2024toolace}. To ensure general abilities, we randomly sampled sentences from the C4 corpus in both English and Japanese, and incorporated the dataset generated by ToolACE to enhance our function-calling capabilities.

\subsubsection{KV Cache Quantization}
Each token processed by the PLaMo 2.0-31B model requires storing 55,296 KV elements. When storing in BF16, this amounts to approximately 110KB per token, a significant memory burden, especially during long-context generation. By enabling static 8-bit KV cache quantization, we were able to reduce this to just 54KB, cutting the per-token KV cache memory footprint in half.
For KV cache quantization, we explored E4M3 and E5M2 floating-point formats \citep{micikevicius2022fp8formatsdeeplearning}. Without any calibration, static quantization using E4M3 slightly outperformed E5M2 across a variety of tasks. This indicates that PLaMo 2.0-31B’s KV activations exhibit a highly stable distribution, making them inherently suitable for quantization.

Thanks to the nature of the static quantization, both the quantization and dequantization procedure of the KV cache incur minimal overhead. Since FlashAttention2~\citep{dao2023flashattention} does not support FP8 computation, we adopted FlashInfer~\citep{ye2025flashinfer} as the backend for attention operations. Compared to xFormers~\citep{xFormers2022}, FlashInfer offers better performance, allowing us to maintain comparable throughput to weight-only quantization while greatly reducing memory footprint.

\begin{table}[t]
\centering
\caption{Jaster benchmark scores for PLaMo 2.0-31B with multiple quantization settings}
\label{jaster-quant}
\scriptsize
\begin{tabular}{lrrrrrrrrrr}
\toprule
Weight/KV Cache & AVG & EL & FA & HE & MC & MR & MT & NLI & QA & RC \\
\midrule
BF16/BF16 & 0.665 & 0.466 & $\mathbf{0.524}$ & $\mathbf{0.645}$ & 0.860 & $\mathbf{0.840}$ & $\mathbf{0.884}$ & 0.852 & \textbf{0.662} & \textbf{0.915} \\
BF16/FP8(E4M3) & 0.670 & $\mathbf{0.578}$  & 0.478 & 0.620 & $\mathbf{0.930}$ & $\mathbf{0.840}$ & 0.862 & $\mathbf{0.856}$ & 0.629 & 0.906 \\
BF16/FP8(E5M2) & 0.659 & 0.484 & 0.475 & 0.630 &$\mathbf{0.930}$ & 0.810 & 0.860 & 0.854 & 0.641 & 0.901 \\
INT4/BF16 & $\mathbf{0.671}$ & 0.550 & 0.489 & 0.620 & $\mathbf{0.930}$ & $\mathbf{0.840}$ & 0.862 & 0.854 & $\mathbf{0.662}$ & 0.901 \\
INT4/FP8(E4M3) & 0.662 & 0.532 & 0.480 & 0.620 & 0.920 & 0.800 & 0.862 & 0.852 & 0.651 & 0.899 \\
INT4/FP8(E5M2) & 0.659 & 0.492 & 0.468 & 0.630 & 0.920 & 0.810 & 0.860 & 0.854 & 0.655 & 0.901 \\
\bottomrule
\end{tabular}
\end{table}

\begin{table}[t]
\centering
\caption{M-IFEval Japanese subset benchmark scores for PLaMo 2.0-31B with multiple quantization settings}
\label{ifeval-ja-quant}
\scriptsize
\begin{tabular}{lrrr}
\toprule
Weight/KV Cache & AVG & Strict Prompt Level & Strict Instruction Level \\
\midrule
BF16/BF16 & $0.677$ & $0.651$ & $\mathbf{0.704}$ \\
BF16/FP8(E4M3) & $0.674$ & $0.645$ & $0.703$\\
BF16/FP8(E5M2) & $0.675$ & $0.645$ & $\mathbf{0.704}$ \\
INT4/BF16 & $0.663$ & $0.640$ & $0.686$ \\
INT4/FP8(E4M3) & $\mathbf{0.684}$ & $\mathbf{0.663}$ & $\mathbf{0.704}$ \\
INT4/FP8(E5M2) & $0.678$ & $0.651$ & $\mathbf{0.704}$ \\
\bottomrule
\end{tabular}
\end{table}

\begin{table}[t]
\centering
\caption{pfgen-bench scores for PLaMo 2.0-31B with multiple quantization settings}
\label{pfgen-quant}
\scriptsize
\begin{tabular}{lrrrrr}
\toprule
Weight/KV Cache & Score & Length & Fluency & Truthfulness & Helpfulness \\
\midrule
BF16/BF16 & $\mathbf{0.890}$ \,($\pm 0.012/\sqrt{10}$) & $104.4$ \,($\pm 11.3$) & $\mathbf{0.951}$ & $0.952$ & $\mathbf{0.767}$ \\
BF16/FP8(E4M3) & $0.8791$ \,($\pm 0.0185/\sqrt{10}$) & $104.2$ \,($\pm 11.3$) & $0.947$ & $\mathbf{0.953}$ & $0.738$ \\
BF16/FP8(E5M2) & $0.8786$ \,($\pm 0.0165/\sqrt{10}$) & $105.1$ \,($\pm 11.2$) & $0.946$ & $0.950$ & $0.740$ \\
INT4/BF16 & $0.885$ \,($\pm 0.0189/\sqrt{10}$) & $104.6$ \,($\pm 10.7$) & $0.944$ & $0.946$ & $0.765$ \\
INT4/FP8(E4M3) & $0.884$ \,($\pm 0.0203/\sqrt{10}$) & $105.0$ \,($\pm 10.9$) & $0.947$ & $ 0.949$ & $0.754$ \\
INT4/FP8(E5M2) & $0.882$ \,($\pm 0.0149/\sqrt{10}$) & $105.1$ \,($\pm 11.1$) & $0.946$ & $0.949$ & $0.750$ \\
\bottomrule
\end{tabular}
\end{table}

\begin{table}[t]
    \centering
    \caption{BFCL V2~\citep{patil2025bfcl} simple scores for PLaMo 2.0-31B with multiple quantization settings}
    \label{bfcl-quant}
    \scriptsize
    \begin{tabular}{lr}
        \toprule
        Weight/KV Cache & Score \\
        \midrule
        BF16/BF16 & $\mathbf{87.50}$\%  \\
        BF16/FP8(E4M3) & $87.00$\% \\
        BF16/FP8(E5M2) & $86.00$\% \\
        INT4/BF16 & $85.75$\%\\
        INT4/FP8(E4M3) & $85.25$\%\\
        INT4/FP8(E5M2) & $84.75$\%\\
        \bottomrule
    \end{tabular}
\end{table}

\subsubsection{Results}

\Cref{jaster-quant,ifeval-ja-quant,pfgen-quant,bfcl-quant} shows the results of our accuracy benchmarks. The leftmost column of the table represents weights and KV Cache dtype, respectively.
In all benchmarks, we observe that 4-bit quantization causes minimal degradation.

At runtime, we leveraged the Marlin~\citep{frantar2025marlin} kernel to accelerate both the prefill and decode stages of inference. 
In our benchmarks, for prefill sequences of up to 4096 tokens, Marlin consistently outperforms the BF16 PyTorch matrix multiplication kernels used by default in vLLM and delivers noticeably faster performance; and for decode sequences of up to 512 tokens, it maintained a clear advantage as well.
We also compared FlashInfer and FlashAttention2 in both stages. The results showed that FlashInfer is better suited for prefill, while FlashAttention2 performs better during decode. Since the PLaMo 2.0-31B model in full precision cannot run on a single L40S GPU, we use the 8B model to present a fair throughput comparison before and after quantization on L40S.

In addition, enabling KV cache quantization helps reduce the occurrence of recomputation under high concurrency. Without quantization, memory insufficiency becomes more pronounced, causing eviction to occur earlier. This in turn triggers recomputation, leading to a significant performance degradation when concurrency exceeds 128. However, with quantization enabled, the threshold increases to 256 concurrent requests, effectively extending the usable range before recomputation is triggered.

\section{Conclusion}

In this paper, we introduced PLaMo 2, a new series of high-performance large language models with a strong focus on Japanese language capabilities and training efficiency. We detailed our comprehensive approach, which integrates architectural innovations, advanced data strategies, and efficient training methodologies. By transitioning from a standard Transformer to a Samba-based architecture and later incorporating full attention, we addressed the dual challenges of inference efficiency and long-context retrieval.

A cornerstone of our work was the extensive use of LLM-generated synthetic data, which enabled us to overcome the scarcity of high-quality Japanese training corpora and significantly boost performance in areas like coding and mathematics. Furthermore, our adoption of weight reusing and pruning techniques allowed for the computationally efficient development of a family of models, demonstrating that a smaller, pruned 8B model could achieve performance on par with a much larger 100B model from a previous generation.

Our evaluation results confirm the success of these strategies. The PLaMo 2 models consistently outperform other open models of similar size on a wide range of Japanese benchmarks, including Jaster, M-IFEval, and pfgen-bench, demonstrating superior instruction-following ability, knowledge of Japan-specific topics, and natural language fluency.

Despite these achievements, our analysis also highlighted areas for future improvement, particularly in complex multi-step and mathematical reasoning, where some models still hold an edge. Future work will focus on enhancing these reasoning capabilities through targeted data curation and more advanced post-training techniques.
We believe the models and methodologies presented here represent a significant contribution to the development of open-source LLMs, particularly for non-English languages, and we will continue to build upon this work to further advance the field.

\section*{Author Contributions}
Within each section, contributors are listed in alphabetical order by last name.

\paragraph{Pre-train}
Yuta Hirokawa,
Kentaro Imajo,
Hiroaki Mikami,
Shogo Murai,
Daisuke Nishino,
Toru Ogawa,
Shuji Suzuki,
Kuniyuki Takahashi,
Tianqi Xu

\paragraph{Post-train}
Kaizaburo Chubachi,
Yasuhiro Fujita,
Toshiki Kataoka,
Goro Kobayashi,
Kento Nozawa,
Shunta Saito,
Avinash Ummadisingu

\paragraph{Inference Performance Optimization}
Shinichi Hemmi, Kenichi Maehashi, Calvin Metzger, Shintarou Okada, Hanqin Wang, Sixue Wang

\paragraph{Overall Project Management}
Daisuke Okanohara, Shotaro Sano, Daisuke Tanaka

\section*{Acknowledgments}
PLaMo 2 models are trained under the project, ``Research and Development Project of the Enhanced Infrastructures for Post 5G Information and Communication System'' (JPNP 20017), subsidized by the New Energy and Industrial Technology Development Organization (NEDO).

We would like to thank PFN members for their helpful discussions and/or implementation support.
We also would like to thank cluster team members for the infrastructure support.

\bibliographystyle{plainnat}
\bibliography{ref} 

\end{document}